\documentclass[12pt]{article}

\usepackage{graphicx}
\usepackage{multirow}
\usepackage{amsmath,amssymb,amsfonts}
\usepackage{amsthm}
\usepackage{mathrsfs}
\usepackage[title]{appendix}
\usepackage{xcolor}
\usepackage{textcomp}
\usepackage{booktabs}
\usepackage{algorithm}
\usepackage{algorithmicx}
\usepackage{algpseudocode}
\usepackage{caption}
\usepackage{bm}
\usepackage{url}
\usepackage[T1]{fontenc}
\usepackage{lmodern}
\usepackage{hyperref}
\usepackage{cleveref}

\begin{document}

\title{Beyond Convolution: A Taxonomy of Structured Operators
for Learning-Based Image Processing}

\begin{center}
{\fontsize{18pt}{22pt}\selectfont\bfseries Beyond Convolution: A Taxonomy of Structured Operators for Learning-Based Image Processing\par}
\vspace{1em}
{\large Simone Cammarasana}
\vspace{0.5em}

{\small CNR-IMATI, Via De Marini 6, Genova, Italy}\\
{\small \texttt{simone.cammarasana@ge.imati.cnr.it}}
\vspace{1.5em}
\end{center}

\begin{abstract}
The convolution operator is the fundamental building block of modern
convolutional neural networks (CNNs), owing to its simplicity,
translational equivariance, and efficient implementation.
However, its structure as a fixed, linear, locally-averaging operator
limits its ability to capture structured signal properties such as
low-rank decompositions, adaptive basis representations, and
non-uniform spatial dependencies.
This paper presents a systematic taxonomy of operators that extend or
replace the standard convolution in learning-based image processing
pipelines.
We organise the landscape of alternative operators into five families:
(i)~\emph{decomposition-based operators}, which separate structural and
noise components through singular value or tensor decompositions;
(ii)~\emph{adaptive weighted operators}, which modulate kernel
contributions as a function of spatial position or signal content;
(iii)~\emph{basis-adaptive operators}, which optimise the analysis
bases together with the network weights;
(iv)~\emph{integral and kernel operators}, which generalise the
convolution to position-dependent and non-linear kernels; and
(v)~\emph{attention-based operators}, which relax the locality
assumption entirely.
For each family, we provide a formal definition, a discussion of its
structural properties with respect to the convolution, and a critical
analysis of the tasks for which the operator is most appropriate.
We further provide a comparative analysis of all families across
relevant dimensions---linearity, locality, equivariance, computational
cost, and suitability for image-to-image and image-to-label tasks---and
outline the open challenges and future directions of this research area.
\end{abstract}


\section{Introduction}
\label{sec:intro}

Deep learning has established itself as the dominant paradigm for
signal and image processing, achieving state-of-the-art performance
on tasks ranging from image restoration~\cite{zhang2017beyond,
dabov2006image} and super-resolution~\cite{dong2015image} to
semantic segmentation and medical image
analysis~\cite{cammarasana2026ultrasound}.
At the heart of nearly every successful architecture lies the
\emph{convolution operator}, which extracts local features by sliding
a fixed, learned kernel over the input signal.
Despite its success, the convolution operator is fundamentally a
\emph{weighted local average}: it applies a fixed linear combination
to each neighbourhood of pixels, sharing the same weights across all
spatial positions.

This structural simplicity, while computationally advantageous, has
well-known limitations.
First, the convolution treats all spatial positions uniformly, making
it insensitive to local signal structure such as edges, textures, or
noise patterns.
Second, its linear nature prevents it from performing operations that
are inherently \emph{structural}, such as separating a signal into
low-rank and high-rank components, or projecting onto an
application-specific basis.
Third, the fixed kernel size imposes a rigid locality prior that is
inappropriate for tasks requiring global context or multi-scale
reasoning.

A growing body of work addresses these limitations by proposing
operators that enrich or replace the convolution while remaining
compatible with the back-propagation training framework.
These contributions are distributed across multiple communities---signal
processing, numerical linear algebra, fuzzy mathematics, and deep
learning---making it difficult to obtain a unified view of the design
space.
Our work fills this gap by providing a \emph{systematic taxonomy} of
structured operators as alternatives or enhancements to convolution
in learning-based image processing.

\paragraph{Motivating examples}
The limitations of standard convolution are well illustrated by
concrete tasks.
In image denoising, applying the singular value decomposition (SVD) to
local patches explicitly separates the structured (low-rank) signal
component from the unstructured noise component~\cite{gu2014weighted,
cammarasana2022denoising,hu2018comprehensive}.
In image super-resolution, interpolating bases such as B-splines
encode smoothness priors that a fixed kernel cannot capture
efficiently~\cite{dong2015image,shi2016real}.
In classification, optimising the density function applied to the
convolution kernel yields a spatially non-uniform weighting that
improves convergence and accuracy~\cite{cammarasana2025weighted,
cammarasana2025classification}.
In fuzzy signal processing, the F-transform with adaptive membership
functions defines a projection operator whose bases are data-driven
rather than fixed~\cite{cammarasana2024ftransform,perfilieva2006fuzzy}.
These examples suggest that the \emph{choice of operator} is not
merely an implementation detail but a fundamental modelling decision
that encodes prior knowledge about the signal and the task.

\paragraph{Contributions}
The main contributions of this paper are the following.
\begin{itemize}
  \item We introduce a principled taxonomy of five families of
  structured operators that extend or replace the convolution in
  learning-based image processing, covering decomposition-based,
  adaptive weighted, basis-adaptive, integral and kernel, and
  attention-based operators.
  \item For each family, we provide a unified formal treatment,
  identifying the structural property of the convolution that each
  operator relaxes or replaces.
  \item We present a comparative analysis of all families across
  multiple dimensions relevant to practitioners, including linearity,
  locality, equivariance, computational cost, and task suitability.
  \item We discuss the open challenges in the area and highlight
  directions for future work, with particular attention to biomedical
  imaging and volumetric data.
\end{itemize}

\paragraph{Relation to own previous work}
Several contributions in this taxonomy build upon and contextualise
our own prior work.
The decomposition-based family is grounded in our learning-based
low-rank denoising approach~\cite{cammarasana2022denoising}, where a
network predicts the optimal SVD thresholds for local image patches.
The adaptive weighted family extends the framework of our weighted
convolution with optimal density
functions~\cite{cammarasana2025weighted,cammarasana2025classification}.
The basis-adaptive family encompasses our work on adaptive membership
functions for the F-transform~\cite{cammarasana2024ftransform}.
The computational aspects of operator optimisation are informed by our
study of high-performance solvers for signal processing
minimisation problems~\cite{cammarasana2025hpc}.
Applications to medical imaging throughout the paper are motivated
by our survey on ultrasound image processing~\cite{cammarasana2026ultrasound}.


\section{Background: The Convolution Operator}
\label{sec:background}

We recall the definition and properties of the discrete convolution
operator as it is used in CNNs, which serves as the common reference
point throughout the taxonomy.

\subsection{Definition}

Given a 2D input signal $\bm{I} \in \mathbb{R}^{R \times C}$ and a
kernel $\bm{w} \in \mathbb{R}^{K_a \times K_b}$, the discrete
convolution is defined as
\begin{equation}
  (\bm{I} \ast \bm{w})_{ij}
  \;:=\;
  \sum_{a=1}^{K_a} \sum_{b=1}^{K_b}
  \bm{w}_{ab} \cdot \bm{I}_{i+a-\lfloor K_a/2\rfloor,\;
                             j+b-\lfloor K_b/2\rfloor},
  \label{eq:conv}
\end{equation}
for $i = 1,\ldots,R$ and $j = 1,\ldots,C$.
In a CNN with $F$ output channels and $D$ input channels, a tensor of
kernels $\bm{W} \in \mathbb{R}^{K_a \times K_b \times D \times F}$
is applied to produce the feature map $\bm{I} \ast \bm{W}$.
The kernel weights are the trainable parameters of the layer,
optimised by minimising a task-specific loss function via
back-propagation~\cite{lecun1989backpropagation,ruder2016overview}.

\subsection{Structural Properties}

The convolution operator has four structural properties that
collectively define its expressive power and its limitations.

\paragraph{Linearity}
The map $\bm{I} \mapsto \bm{I} \ast \bm{w}$ is linear in both
$\bm{I}$ and $\bm{w}$.
This enables efficient gradient computation but prevents the operator
from modelling non-linear local interactions.

\paragraph{Translational equivariance}
The same kernel $\bm{w}$ is applied at every spatial position, so
that shifting the input produces a correspondingly shifted output:
$T_\delta(\bm{I}) \ast \bm{w} = T_\delta(\bm{I} \ast \bm{w})$,
where $T_\delta$ denotes a spatial shift by $\delta$.
This inductive bias is appropriate for natural images but may be
suboptimal for structured data with position-dependent statistics,
such as medical images with fixed anatomical priors.

\paragraph{Locality}
The output at position $(i,j)$ depends only on the $K_a \times K_b$
neighbourhood $\mathcal{N}(\bm{I}_{ij})$.
This limits the receptive field of each layer and requires deep
stacking of layers to capture long-range dependencies.

\paragraph{Uniform weighting}
The kernel applies the same learned weight to each relative position
in the neighbourhood, irrespective of the local content of the
signal.
In particular, the kernel does not distinguish between positions that
carry structural information (e.g., edges) and positions that carry
noise.

\subsection{Computational Cost}

For an input of size $R \times C$, $F$ output channels, and kernel
size $K \times K$, the computational cost of evaluating
Eq.~\eqref{eq:conv} is $\mathcal{O}(RCF K^2)$.
Practical implementations exploit FFT-based
convolution~\cite{mathieu2013fast}, the Winograd
algorithm~\cite{lavin2016fast}, and GEMM-based
unrolling~\cite{chetlur2014cudnn} to reduce the constant factor,
while maintaining the same asymptotic complexity.

\subsection{The Convolution as a Reference Point}

Table~\ref{tab:conv_properties} summarises the four structural
properties discussed above.
Each operator family introduced in
Sections~\ref{sec:decomp}--\ref{sec:attention} relaxes or replaces
one or more of these properties in a principled way, motivated by the
structure of the signal and the requirements of the task.
This framing provides the organising principle of the taxonomy.

\begin{table}[t]
\centering
\caption{Structural properties of the standard convolution operator,
used as a reference point for the taxonomy.}
\label{tab:conv_properties}
\begin{tabular}{lll}
\toprule
\textbf{Property} & \textbf{Convolution} & \textbf{Implication} \\
\midrule
Linearity          & Yes & Efficient gradient; no local non-linearity \\
Transl.\ equivariance & Yes & Position-independent feature extraction \\
Locality           & Yes (fixed $K$) & Limited receptive field per layer \\
Uniform weighting  & Yes & Content-agnostic neighbourhood averaging \\
\bottomrule
\end{tabular}
\end{table}

\section{Decomposition-Based Operators}
\label{sec:decomp}

Decomposition-based operators replace the convolution's uniform
averaging with a factorisation that explicitly separates the signal
into structural components.
The canonical example is the Singular Value Decomposition (SVD), which
decomposes a matrix into orthogonal bases weighted by singular values,
providing a natural separation between high-energy (structural) and
low-energy (noise) components.

\subsection{Local SVD Operators}

Given a patch $\bm{P} \in \mathbb{R}^{m \times n}$ extracted from
the neighbourhood of a pixel, the thin SVD is
$\bm{P} = \bm{U} \bm{S} \bm{V}^\top$,
where $\bm{U} \in \mathbb{R}^{m \times r}$,
$\bm{S} = \mathrm{diag}(s_1,\ldots,s_r)$,
$\bm{V} \in \mathbb{R}^{n \times r}$, and $r = \min(m,n)$.
A denoised approximation is obtained by applying a
threshold $\tau$ to the singular values:
\begin{equation}
  \hat{\bm{P}} = \bm{U}\,\mathrm{diag}(\tau(\bm{s}))\,\bm{V}^\top,
  \label{eq:svd_thresh}
\end{equation}
where $\tau(\bm{s}) = \max(\bm{s} - \bm{\lambda}, \bm{0})$
for a vector of thresholds $\bm{\lambda}$.
This operation is \emph{non-linear} (the threshold is a non-linear
function of the singular values) and \emph{structure-aware} (it
explicitly separates the low-rank signal from the residual noise
component).

The Weighted Nuclear Norm Minimisation
(WNNM)~\cite{gu2014weighted} applies this principle by combining SVD
thresholding with block-matching to aggregate similar patches into a
3D stack, a strategy also used in BM3D~\cite{dabov2006image}.
Our learning-based extension~\cite{cammarasana2022denoising} replaces
the hand-crafted threshold selection with a network that
\emph{predicts} the optimal thresholds $\bm{\lambda}$ from the
singular values of each block, trained to minimise the
reconstruction error with respect to the ground-truth image.
Algorithm~\ref{alg:svd_denoise} summarises this approach.

\begin{algorithm}[t]
\caption{Learning-based SVD denoising~\cite{cammarasana2022denoising}}
\label{alg:svd_denoise}
\begin{algorithmic}[1]
\Require Noisy image $\bm{Y}$, trained network $\mathcal{N}$,
         number of iterations $T$
\Ensure Denoised image $\hat{\bm{X}}$
\State $\hat{\bm{X}}^{(0)} \leftarrow \bm{Y}$
\For{$t = 1, \ldots, T$}
  \State Extract 3D blocks $\{\bm{B}_k\}$ via block matching on
         $\hat{\bm{X}}^{(t-1)}$
  \For{each block $\bm{B}_k$}
    \State Compute $\bm{U}_k, \bm{S}_k, \bm{V}_k \leftarrow
           \mathrm{SVD}(\bm{B}_k)$
    \State Predict thresholds $\hat{\bm{\lambda}}_k \leftarrow
           \mathcal{N}(\mathrm{diag}(\bm{S}_k))$
    \State $\hat{\bm{B}}_k \leftarrow \bm{U}_k\,
           \mathrm{diag}(\bm{S}_k - \hat{\bm{\lambda}}_k)\,
           \bm{V}_k^\top$
  \EndFor
  \State $\hat{\bm{X}}^{(t)} \leftarrow
         \mathrm{Aggregate}(\{\hat{\bm{B}}_k\})$
\EndFor
\State \Return $\hat{\bm{X}}^{(T)}$
\end{algorithmic}
\end{algorithm}

\subsection{Tensor Decomposition Operators}

For volumetric data and multi-channel images, the Higher-Order SVD
(HOSVD)~\cite{kolda2009tensor} generalises the matrix SVD to tensors
$\mathcal{X} \in \mathbb{R}^{n_1 \times \cdots \times n_d}$ via the
Tucker decomposition
$\mathcal{X} \approx \mathcal{G} \times_1 \bm{U}^{(1)}
 \times_2 \bm{U}^{(2)} \cdots \times_d \bm{U}^{(d)}$,
where $\mathcal{G}$ is the core tensor and $\bm{U}^{(k)}$ are mode-$k$
factor matrices.
Applied locally to image patches, HOSVD simultaneously exploits
spatial, channel, and depth correlations, making it particularly
effective for hyperspectral and
volumetric data \cite{zhang2019hyperspectral} .

\subsection{Low-Rank Layer Approximations}

Decomposition principles can also be applied at the \emph{layer} level
rather than the patch level.
Low-rank approximations of the weight
tensor~\cite{sainath2013low,lebedev2014speeding} factorise
$\bm{W} \approx \bm{W}_1 \bm{W}_2$ to reduce the number of
parameters and the computational cost.
While this is primarily a compression technique, it encodes a
structural prior (low-rank kernel) that can improve generalisation
in data-scarce regimes.

\subsection{Properties and Task Suitability}

Decomposition-based operators break the \emph{uniform weighting}
property of the convolution by assigning different contributions to
different spectral components.
They are non-linear (due to thresholding) and content-adaptive (the
decomposition depends on the local patch structure).
They are most naturally suited to \emph{image-to-image tasks} where
noise removal, compression, or structured feature extraction is
required~\cite{hu2018comprehensive}.
Their main limitation is computational cost: computing the SVD of
each patch at every layer is significantly more expensive than a
standard convolution.

\section{Adaptive Weighted Operators}
\label{sec:adaptive}

Adaptive weighted operators retain the local neighbourhood structure
of the convolution but modulate the kernel weights as a function of
spatial position, signal content, or an externally optimised density
function.
They relax the \emph{uniform weighting} and, in some cases, the
\emph{translational equivariance} property.

\subsection{Convolution with Density Functions}

A density function $\bm{\Phi} \in \mathbb{R}^{K \times K}$ scales the
kernel weights element-wise before applying the convolution.
Given the standard kernel $\bm{w}$ and the density function
$\bm{\Phi}$, the weighted convolution is defined
as~\cite{cammarasana2025weighted}
\begin{equation}
  (\bm{I} \ast \bm{w}_{\bm{\Phi}})_{ij}
  \;:=\;
  \sum_{a,b} (\bm{\Phi}_{ab}\,\bm{w}_{ab})\,
  \bm{I}_{i+a-\lfloor K/2\rfloor,\,j+b-\lfloor K/2\rfloor}.
  \label{eq:wconv}
\end{equation}
The density function encodes a prior about the relative importance of
pixels at different distances from the reference pixel.
When $\bm{\Phi} = \bm{1}$, Eq.~\eqref{eq:wconv} reduces to the
standard convolution.

The optimal density function is computed by solving an outer
optimisation problem~\cite{cammarasana2025weighted}:
given a learning model $\mathcal{M}_{\bm{\Phi}}$ that minimises the
loss function $\mathcal{L}$ with respect to the kernel weights
$\bm{W}$ for a fixed $\bm{\Phi}$, the optimal density function
$\bm{\Phi}^*$ is found by minimising the value of this minimised
loss over all admissible density functions:
\begin{equation}
  \bm{\Phi}^* = \arg\min_{\bm{\Phi}}\;
  \min_{\bm{W}} \mathcal{L}(\mathcal{T},\,
  \hat{\mathcal{T}}(\bm{W}_{\bm{\Phi}})).
  \label{eq:outer}
\end{equation}
Crucially, this formulation \emph{separates} the optimisation of the
kernel weights (solved by stochastic gradient descent) from the
optimisation of the density function (solved by a global
derivative-free method such as DIRECT-L~\cite{cammarasana2025hpc,
gablonsky2000locally}).
Algorithm~\ref{alg:density} outlines the procedure.

\begin{algorithm}[t]
\caption{Optimal density function
computation~\cite{cammarasana2025weighted}}
\label{alg:density}
\begin{algorithmic}[1]
\Require Dataset $(\mathcal{I}, \mathcal{T})$, kernel size $K$,
         bounds $[\alpha_{\min}, \alpha_{\max}]$
\Ensure Optimal density function $\bm{\Phi}^*$
\State Initialise $\bm{\alpha} = [1, \ldots, 1]$
\Repeat
  \State $\bm{\Phi} \leftarrow \bm{\alpha}\bm{\alpha}^\top$
  \State $\bm{W}^* \leftarrow \arg\min_{\bm{W}}
         \mathcal{L}(\mathcal{T}, \hat{\mathcal{T}}(\bm{W}_{\bm{\Phi}}))$
         \hfill \textit{(SGD)}
  \State Evaluate $f(\bm{\alpha}) \leftarrow
         \mathcal{L}(\mathcal{T}, \hat{\mathcal{T}}(\bm{W}^*_{\bm{\Phi}}))$
  \State Update $\bm{\alpha}$ via DIRECT-L using $f(\bm{\alpha})$
\Until{convergence}
\State \Return $\bm{\Phi}^* \leftarrow \bm{\alpha}^*(\bm{\alpha}^*)^\top$
\end{algorithmic}
\end{algorithm}

\subsection{Dynamic Convolution}

Dynamic convolution~\cite{chen2020dynamic} aggregates multiple
parallel kernels $\{\bm{w}^{(k)}\}_{k=1}^K$ with input-dependent
attention weights $\pi_k(\bm{I})$:
\begin{equation}
  \bm{I} \ast \bm{w}_{\mathrm{dyn}}
  \;:=\;
  \bm{I} \ast \left(\sum_{k=1}^K \pi_k(\bm{I})\,\bm{w}^{(k)}\right).
\end{equation}
The attention weights $\pi_k$ are produced by a small auxiliary
network conditioned on a global average-pooled representation of the
input.
This makes the effective kernel content-dependent while preserving the
local structure and translational equivariance of the convolution.
Omni-dimensional dynamic convolution~\cite{li2022omni} extends this
idea to multi-dimensional attention over spatial, channel, and filter
dimensions simultaneously.

\subsection{Deformable Convolution}

Deformable convolution~\cite{dai2017deformable} learns a set of
spatial offsets $\{\Delta p_k\}$ that shift the sampling locations of
the kernel:
\begin{equation}
  (\bm{I} \ast_{\mathrm{def}} \bm{w})_{ij}
  \;:=\;
  \sum_{k} \bm{w}_k \cdot \bm{I}(p_{ij} + p_k + \Delta p_k),
\end{equation}
where bilinear interpolation is used for non-integer sampling
positions.
Deformable convolution relaxes both the \emph{uniform weighting} and
the \emph{locality} property: the receptive field adapts to the
geometric structure of the signal.

\subsection{Properties and Task Suitability}

Adaptive weighted operators are particularly effective in scenarios
where the relative importance of neighbouring pixels varies across
the image, such as tasks involving structured textures, edges, or
anisotropic noise.
Density function optimisation~\cite{cammarasana2025weighted,
cammarasana2025classification} improves convergence speed and
final accuracy for both image-to-image (denoising, PSNR improvement
of 6--7\%) and image-to-label (classification, accuracy improvement
of 7 percentage points) tasks, without increasing the number of
trainable parameters.
The computational overhead is modest ($\approx 7$\% on GPU).

\section{Basis-Adaptive Operators}
\label{sec:basis}

Basis-adaptive operators define the analysis and synthesis bases as
learnable or data-dependent objects, replacing the fixed Fourier-like
bases implicit in the standard convolution.

\subsection{F-Transform with Adaptive Membership Functions}

The \emph{fuzzy transform} (F-transform)~\cite{perfilieva2006fuzzy}
projects a signal $f$ onto a set of fuzzy partition functions
$\{A_k\}_{k=1}^n$ defined over the signal domain.
The $k$-th F-transform component is:
\begin{equation}
  F_k
  \;:=\;
  \frac{\sum_{x \in \Omega} f(x)\,A_k(x)}
       {\sum_{x \in \Omega} A_k(x)}.
\end{equation}
The standard F-transform uses fixed, regularly spaced membership
functions.
Our extension~\cite{cammarasana2024ftransform} optimises the
membership functions $\{A_k\}$ jointly with the network weights,
making the projection bases \emph{adaptive} to the signal statistics.
Algorithm~\ref{alg:ftransform} summarises the adaptive F-transform
layer.

\begin{algorithm}[t]
\caption{Adaptive F-transform layer~\cite{cammarasana2024ftransform}}
\label{alg:ftransform}
\begin{algorithmic}[1]
\Require Input signal $f$, initialised membership functions
         $\{A_k^{(0)}\}$, loss $\mathcal{L}$
\Ensure Adapted membership functions $\{A_k^*\}$, output components
        $\{F_k\}$
\Repeat
  \For{each $k$}
    \State $F_k \leftarrow
           \sum_x f(x)\,A_k(x) / \sum_x A_k(x)$
  \EndFor
  \State Compute $\mathcal{L}$; backpropagate gradients w.r.t.\
         $\{A_k\}$ and network weights
  \State Update $\{A_k\}$ and network weights via SGD
\Until{convergence}
\State \Return $\{A_k^*\}$, $\{F_k\}$
\end{algorithmic}
\end{algorithm}

The F-transform provides a natural connection between fuzzy set theory
and signal processing, allowing the incorporation of domain knowledge
into the choice and shape of the partition functions.
Unlike the convolution, which applies the same kernel at every
position, the F-transform aggregates information over the entire
domain with position-dependent weights, relaxing both the locality
and the translational equivariance constraints.

\subsection{Learnable Wavelet Transforms}

Wavelets provide a multi-scale decomposition of the signal, capturing
features at different resolutions.
Learnable wavelet layers~\cite{liu2019multi} parameterise the wavelet
filters and optimise them end-to-end, combining the interpretability
of wavelet theory with the flexibility of learned
representations~\cite{rakheja2017image}.
The \emph{shearlet transform}~\cite{kutyniok2016shearlab} extends
wavelets with directional sensitivity, providing a richer basis for
capturing anisotropic features such as edges and ridges.
Applied to denoising, shearlets are particularly effective for images
with strong directional structures.

\subsection{Sparse Dictionary Learning}

An over-complete dictionary $\bm{D} \in \mathbb{R}^{n \times M}$
($M \gg n$) represents the signal as a sparse linear combination of
atoms: $\bm{y} \approx \bm{D}\bm{a}$, $\|\bm{a}\|_0 \leq s$.
The K-SVD algorithm~\cite{aharon2006k} alternates between sparse
coding (finding $\bm{a}$ given $\bm{D}$) and dictionary update
(refining each atom via SVD of the representation error).
Dictionary learning can be embedded into a CNN as a learnable layer,
replacing the convolution with a pursuit-and-synthesis operation.

\subsection{Properties and Task Suitability}

Basis-adaptive operators relax the \emph{translational equivariance}
and \emph{uniform weighting} properties.
Their strength lies in tasks where the signal has a known or learnable
structure in a transformed domain (e.g., sparsity in a wavelet
basis, smoothness in a fuzzy partition).
They are particularly useful in medical imaging, where physical
acquisition models (e.g., ultrasound speckle, MRI $k$-space
sampling) motivate specific choices of
basis~\cite{cammarasana2026ultrasound}.

\section{Integral and Kernel Operators}
\label{sec:integral}

Integral and kernel operators generalise the convolution by allowing
the kernel to depend on the absolute or relative position of the
pixels, rather than only on their relative offset.
This relaxes the \emph{translational equivariance} property.

\subsection{Non-Local Means and Non-Local Neural Networks}

The Non-Local Means (NLM) filter~\cite{buades2005non} computes the
output at position $i$ as a weighted average over \emph{all}
positions $j$:
\begin{equation}
  \hat{f}(i) = \frac{1}{Z(i)}\sum_j w(i,j)\,f(j),
  \quad
  w(i,j) = \exp\!\left(-\frac{\|\bm{P}_i - \bm{P}_j\|^2}{h^2}\right),
  \label{eq:nlm}
\end{equation}
where $\bm{P}_k$ is the patch at position $k$ and $h$ is a
bandwidth parameter.
The weight $w(i,j)$ depends on the similarity between patches at $i$
and $j$, making the operator content-adaptive and non-local.
Non-Local Neural Networks~\cite{wang2018non} embed this idea into a
differentiable layer, computing pairwise affinities between all
feature map positions.

\subsection{Radial Basis Function Networks}

Radial Basis Function (RBF) networks~\cite{broomhead1988radial}
define the output as a linear combination of radially symmetric basis
functions:
\begin{equation}
  y(\bm{x}) = \sum_{k=1}^M c_k\,\phi(\|\bm{x} - \bm{\mu}_k\|),
\end{equation}
where $\phi$ is a radially symmetric function (e.g., Gaussian,
multiquadric), $\bm{\mu}_k$ are the centres, and $c_k$ are the
coefficients.
Adaptive kernel-based sampling~\cite{cammarasana2021kernel} optimises the centres, widths, and coefficients of Gaussian kernels to approximate arbitrary signals, including 2D images and vector fields, providing a structured alternative to fixed-grid convolution.
Applied to image super-resolution, RBF interpolation with
data-dependent centres provides a structured alternative to
pixel-shuffle upsampling, encoding smoothness priors that are not
available to a standard transposed convolution~\cite{dong2015image}.

\subsection{Convolutional Kernel Networks}

Convolutional Kernel Networks (CKNs)~\cite{mairal2014convolutional}
replace the dot product in the convolution with a positive definite
kernel function $k(\cdot, \cdot)$:
\begin{equation}
  (\bm{I} \ast_k \bm{w})_{ij}
  \;:=\;
  \sum_{a,b} k(\mathcal{N}(\bm{I}_{ij}),\,\bm{w}_{ab}),
\end{equation}
where the kernel $k$ is typically a Gaussian or polynomial function
over patch space.
CKNs provide a principled connection between deep learning and
kernel methods~\cite{scholkopf2002learning}, and can be extended to
graph-structured
signals~\cite{chen2020convolutional}.

\subsection{Position-Encoding Operators}

Coordinate convolution (CoordConv)~\cite{liu2018intriguing} augments
the input feature map with explicit coordinate
channels, allowing the network to break translational equivariance
in a controlled way.
This is particularly useful for tasks where the absolute position of
features is informative, such as object detection and localisation.

\subsection{Properties and Task Suitability}

Integral and kernel operators are the most general family in the
taxonomy.
They can model arbitrary dependencies between input positions, at
the cost of increased computational complexity (NLM requires
$\mathcal{O}(N^2)$ operations per pixel for a full non-local
search window) and reduced inductive bias.
They are most effective in tasks requiring long-range
context~\cite{wang2018non} or tasks where the translational
equivariance prior is inappropriate.
A key challenge is efficient implementation on large
inputs~\cite{cammarasana2025hpc}.

\section{Attention-Based Operators}
\label{sec:attention}

Attention-based operators can be viewed as the extreme case of the
integral operator family, where the kernel is entirely learned from
the data and depends on the global content of the input.
They relax all four structural properties of the convolution and have
become the dominant operator in large-scale vision
models~\cite{dosovitskiy2020image,vaswani2017attention}.

\subsection{Self-Attention}

The scaled dot-product self-attention
mechanism~\cite{vaswani2017attention} computes:
\begin{equation}
  \mathrm{Attention}(\bm{Q}, \bm{K}, \bm{V})
  \;=\;
  \mathrm{softmax}\!\left(\frac{\bm{Q}\bm{K}^\top}{\sqrt{d_k}}\right)
  \bm{V},
  \label{eq:attn}
\end{equation}
where the query $\bm{Q}$, key $\bm{K}$, and value $\bm{V}$ matrices
are linear projections of the input features.
Each output position attends to all input positions, with weights
determined by the similarity between query and key vectors.
Self-attention is not translational equivariant, not local, and not
linear in the input (due to the softmax).

\subsection{Spatial and Channel Attention}

Spatial attention~\cite{woo2018cbam} learns a weight map over the
spatial dimensions of the feature map, emphasising task-relevant
regions.
Channel attention~\cite{wang2020eca} learns weights over the channel
dimension, selecting the most informative feature channels.
Both can be applied as lightweight modules that augment a standard
convolutional backbone without replacing it entirely.

\subsection{Vision Transformers}

Vision Transformers (ViT)~\cite{dosovitskiy2020image} replace
convolutional layers entirely with multi-head self-attention applied
to non-overlapping patches (tokens).
This architecture achieves state-of-the-art performance on large
datasets but requires substantially more training data than CNNs due
to its weaker inductive biases.
NAFNet~\cite{chen2022simple} and similar architectures apply
simplified attention mechanisms that reduce the computational cost of
the softmax operation for image restoration tasks.

\subsection{Attention vs.\ Structured Operators}

While attention mechanisms achieve remarkable performance, their
success comes with the cost of weak structural priors and high
computational cost ($\mathcal{O}(N^2)$ in sequence length).
In contrast, the structured operators discussed in
Sections~\ref{sec:decomp}--\ref{sec:integral} embed domain knowledge
into the operator design, which can be beneficial in data-scarce
regimes such as medical imaging, or when interpretability is
required~\cite{cammarasana2026ultrasound}.
Importantly, attention and structured operators are not mutually
exclusive: hybrid architectures that combine local structured
operators with global attention modules represent a promising
research direction.

\section{Comparative Analysis}
\label{sec:comparison}

\subsection{Structural Properties}

Table~\ref{tab:comparison} summarises the structural properties of
all five operator families with respect to the four reference
properties of the standard convolution.

\begin{table*}[t]
\centering
\caption{Comparison of operator families across structural properties
and task suitability.
\textbf{I2I}: image-to-image tasks (denoising, super-resolution).
\textbf{I2L}: image-to-label tasks (classification, detection).
$\circ$: partial; \checkmark: yes; $\times$: no.}
\label{tab:comparison}
\resizebox{\textwidth}{!}{%
\begin{tabular}{lccccccccc}
\toprule
\textbf{Operator family}
  & \textbf{Linear}
  & \textbf{Transl.\ equiv.}
  & \textbf{Local}
  & \textbf{Unif.\ weight.}
  & \textbf{Adapt.\ to content}
  & \textbf{Comp.\ cost}
  & \textbf{I2I}
  & \textbf{I2L}
  & \textbf{Key ref.} \\
\midrule
Standard convolution
  & \checkmark & \checkmark & \checkmark & \checkmark
  & $\times$ & $\mathcal{O}(K^2)$ & \checkmark & \checkmark
  & \cite{lecun1989backpropagation} \\
Decomp.-based (SVD)
  & $\times$ & $\times$ & \checkmark & $\times$
  & \checkmark & $\mathcal{O}(K^3)$ & \checkmark & $\circ$
  & \cite{gu2014weighted,cammarasana2022denoising} \\
Adapt.\ weighted (density)
  & \checkmark & $\circ$ & \checkmark & $\times$
  & $\circ$ & $\mathcal{O}(K^2)$ & \checkmark & \checkmark
  & \cite{cammarasana2025weighted} \\
Adapt.\ weighted (dynamic)
  & \checkmark & \checkmark & \checkmark & $\times$
  & \checkmark & $\mathcal{O}(K^2)$ & \checkmark & \checkmark
  & \cite{chen2020dynamic} \\
Basis-adaptive (F-transform)
  & \checkmark & $\times$ & $\times$ & $\times$
  & \checkmark & $\mathcal{O}(n \cdot |\Omega|)$ & \checkmark & $\circ$
  & \cite{cammarasana2024ftransform,perfilieva2006fuzzy} \\
Basis-adaptive (wavelet)
  & \checkmark & $\times$ & $\circ$ & $\times$
  & $\circ$ & $\mathcal{O}(N \log N)$ & \checkmark & $\circ$
  & \cite{kutyniok2016shearlab} \\
Integral (NLM / non-local)
  & \checkmark & $\times$ & $\times$ & $\times$
  & \checkmark & $\mathcal{O}(N^2)$ & \checkmark & $\circ$
  & \cite{buades2005non,wang2018non} \\
Integral (RBF / kernel)
  & \checkmark & $\times$ & $\times$ & $\times$
  & \checkmark & $\mathcal{O}(M \cdot N)$ & \checkmark & \checkmark
  & \cite{mairal2014convolutional} \\
Attention (self-attention)
  & $\times$ & $\times$ & $\times$ & $\times$
  & \checkmark & $\mathcal{O}(N^2)$ & \checkmark & \checkmark
  & \cite{vaswani2017attention} \\
\bottomrule
\end{tabular}}
\end{table*}

\subsection{Design Principle Perspective}

Viewing the operators through the lens of the four structural
properties reveals a clear design trade-off.
Moving from the standard convolution toward attention-based operators,
each property is progressively relaxed: uniform weighting first
(adaptive weighted operators), then translational equivariance
(basis-adaptive and integral operators), then locality
(attention-based operators).
This relaxation is accompanied by an increase in expressive power and
a corresponding increase in computational cost and reduction in
inductive bias.

\paragraph{The role of inductive bias}
Strong inductive biases---such as locality, translational equivariance,
and uniform weighting---reduce the number of parameters required to
learn a good representation and improve generalisation in
data-scarce regimes.
Weak inductive biases, conversely, allow the model to learn arbitrary
dependencies from large datasets.
The appropriate choice of operator therefore depends on the
availability of training data, the structural properties of the
signal, and the requirements of the task.

\paragraph{Computational cost}
As shown in Table~\ref{tab:comparison}, operators that relax the
locality property (integral operators, attention) have a
significantly higher computational cost than local operators.
For practical deployment on large images or volumetric data, the
efficiency of the operator is a critical
consideration~\cite{cammarasana2025hpc}.
Optimised GPU implementations and approximations (e.g., linear
attention~\cite{katharopoulos2020transformers}) can reduce this cost,
but at the expense of expressive power.

\subsection{Task Suitability Summary}

For \emph{image-to-image} tasks, decomposition-based and
basis-adaptive operators are particularly well-suited because they
explicitly encode structural properties of natural images (low-rank
structure, multi-scale sparsity) that are relevant to denoising and
super-resolution.
For \emph{image-to-label} tasks, adaptive weighted and attention-based
operators are more appropriate, as they can capture global contextual
information that is relevant for recognition and
classification~\cite{cammarasana2025classification}.
In both cases, the optimal choice of operator depends on the specific
signal statistics, the available training data, and the computational
budget.

\section{Open Challenges and Future Directions}
\label{sec:challenges}

\paragraph{Combination of operators}
Most existing work considers operators in isolation.
A promising direction is the design of architectures that combine
multiple operator families, exploiting their complementary strengths.
For example, a decomposition-based layer at the input could remove
noise before a standard convolutional backbone extracts features, or
an attention module could complement local structured operators with
global context.

\paragraph{Operator selection as a meta-learning problem}
Given the diversity of operators and the task-dependence of their
performance, a natural question is whether the choice of operator can
itself be automated.
Neural architecture search (NAS)~\cite{zoph2016neural} techniques
could be extended to search over operator families, not just
architectural hyperparameters.

\paragraph{Extension to 3D and volumetric data}
All operators discussed in this paper are presented in 2D for
clarity, but most have natural 3D extensions.
For volumetric medical data (CT, MRI, ultrasound), structured
operators that exploit the anisotropy of the acquisition process
(e.g., different in-plane vs.\ through-plane resolution) could
provide significant advantages~\cite{cammarasana2026ultrasound}.
Preliminary work on 3D weighted convolution is available
at~\cite{cammarasana2025weighted}.

\paragraph{Interpretability and theoretical analysis}
While the empirical performance of many operators has been
demonstrated, their theoretical properties---approximation power,
convergence of optimisation, generalisation bounds---are less well
understood.
Formal analysis of the optimisation landscape of the density function
optimisation problem~\cite{cammarasana2025weighted} and the
convergence of adaptive F-transform
layers~\cite{cammarasana2024ftransform} are examples of the type of
theoretical grounding that is needed across all operator families.
Efficient solver design for the related minimisation problems also
remains an open
challenge~\cite{cammarasana2025hpc}.

\paragraph{Biomedical and real-world applications}
Medical imaging represents a particularly important application
domain for structured operators, due to the structured noise models
(speckle in ultrasound, Rician noise in MRI), the anisotropic
acquisition geometry, and the frequent data
scarcity~\cite{cammarasana2026ultrasound}.
The operator families discussed in this survey, particularly
decomposition-based and basis-adaptive operators, are well aligned
with these challenges and deserve further investigation in this
context.

\paragraph{Hardware-aware operator design}
The theoretical $50\%$ overhead of the weighted convolution is
reduced to approximately $7\%$ on modern GPU hardware due to
memory-level
parallelism~\cite{cammarasana2025weighted}.
More broadly, the computational properties of operators are
increasingly architecture-dependent, motivating the design of
operators that are co-optimised with the underlying hardware.

\section{Conclusions}
\label{sec:conclusion}

We have presented a systematic taxonomy of structured operators that
extend or replace the standard convolution in learning-based image
processing.
The five families---decomposition-based, adaptive weighted,
basis-adaptive, integral and kernel, and attention-based
operators---differ in which structural properties of the convolution
they relax, and are consequently suited to different signal types,
task requirements, and computational constraints.

The central message of this survey is that the convolution, while
highly effective in many settings, is not the only or always the
optimal choice of operator for learning-based image processing.
A principled selection of the operator family, guided by the
structural properties of the signal and the requirements of the task,
can lead to significant improvements in both accuracy and efficiency.
We hope that this taxonomy provides a useful reference for researchers
and practitioners in the field, and motivates further work on the
design, analysis, and application of structured operators for deep
learning.

\paragraph{Acknowledgements}
SC is part of RAISE Innovation Ecosystem, funded by
the European Union -- NextGenerationEU and by the Ministry of
University and Research (MUR), National Recovery and Resilience Plan,
Mission 4, Component 2, Investment 1.5, project
\emph{``RAISE -- Robotics and AI for Socio-economic Empowerment''}
(ECS00000035).

\section*{Declarations}
\noindent\textbf{Competing interests}
The author declares no competing interests.

\noindent\textbf{Availability of data and materials}
No datasets were generated or analysed during the current study.

\bibliographystyle{alpha}
\bibliography{refs_beyond}

@article{cammarasana2022denoising,
  author    = {Cammarasana, Simone and Patan{\`e}, Giuseppe},
  title     = {Learning-based low-rank denoising},
  journal   = {Signal, Image and Video Processing},
  pages     = {1--7},
  year      = {2022},
  publisher = {Springer},
  doi       = {10.1007/s11760-022-02258-4}
}

@article{cammarasana2025weighted,
  title={Optimal Density Functions for Weighted Convolution in Learning Models},
  author={Cammarasana, Simone and Patan{\`e}, Giuseppe},
  journal={arXiv preprint arXiv:2505.24527},
  year={2025}
}

@article{cammarasana2024ftransform,
  author  = {Cammarasana, Simone and Patan{\`e}, Giuseppe},
  title   = {Adaptive membership functions and {F}-transform},
  journal = {IEEE Transactions on Fuzzy Systems},
  volume  = {32},
  number  = {5},
  pages   = {2786--2796},
  year    = {2024},
  doi     = {10.1109/TFUZZ.2024.3360633}
}

@article{cammarasana2025hpc,
  author  = {Cammarasana, Simone and Patan{\`e}, Giuseppe},
  title   = {Analysis and comparison of high-performance computing solvers
             for minimisation problems in signal processing},
  journal = {Mathematics and Computers in Simulation},
  volume  = {229},
  pages   = {525--538},
  year    = {2025},
  doi     = {10.1016/j.matcom.2024.10.003}
}

@book{cammarasana2026ultrasound,
  author    = {Cammarasana, Simone and Patan{\`e}, Giuseppe},
  title     = {Real-Time Processing of Ultrasound Images:
               State-of-the-Art and Future Perspectives},
  series    = {Synthesis Lectures on Visual Computing},
  publisher = {Springer},
  year      = {2026},
  isbn      = {978-3-032-05804-1},
  doi       = {10.1007/978-3-032-05805-8}
}

@misc{cammarasana2025classification,
  author        = {Cammarasana, Simone and Patan{\`e}, Giuseppe},
  title         = {Optimal Weighted Convolution for Classification and Denoising},
  howpublished  = {arXiv:2505.24558},
  year          = {2025}
}

@article{lecun1989backpropagation,
  author  = {LeCun, Yann and Boser, Bernhard and Denker, John and Henderson, Donnie
             and Howard, Richard and Hubbard, Wayne and Jackel, Lawrence},
  title   = {Backpropagation applied to handwritten zip code recognition},
  journal = {Neural Computation},
  volume  = {1},
  number  = {4},
  pages   = {541--551},
  year    = {1989}
}

@article{ruder2016overview,
  author  = {Ruder, Sebastian},
  title   = {An overview of gradient descent optimisation algorithms},
  journal = {arXiv preprint arXiv:1609.04747},
  year    = {2016}
}

@inproceedings{lavin2016fast,
  author    = {Lavin, Andrew and Gray, Scott},
  title     = {Fast algorithms for convolutional neural networks},
  booktitle = {CVPR},
  pages     = {4013--4021},
  year      = {2016}
}

@article{mathieu2013fast,
  author  = {Mathieu, Micha{\"e}l and Henaff, Mikael and LeCun, Yann},
  title   = {Fast training of convolutional networks through FFTs},
  journal = {arXiv preprint arXiv:1312.5851},
  year    = {2013}
}

@article{chetlur2014cudnn,
  author  = {Chetlur, Sharan and others},
  title   = {cuDNN: Efficient primitives for deep learning},
  journal = {arXiv preprint arXiv:1410.0759},
  year    = {2014}
}

@inproceedings{dabov2006image,
  author    = {Dabov, Kostadin and Foi, Alessandro and Katkovnik, Vladimir and Egiazarian, Karen},
  title     = {Image denoising by sparse 3D transform-domain collaborative filtering},
  booktitle = {SPIE Electronic Imaging},
  year      = {2006}
}

@inproceedings{gu2014weighted,
  author    = {Gu, Shuhang and Zhang, Lei and Zuo, Wangmeng and Feng, Xiangchu},
  title     = {Weighted nuclear norm minimisation with application to image denoising},
  booktitle = {CVPR},
  pages     = {2862--2869},
  year      = {2014}
}

@article{zhang2017beyond,
  author  = {Zhang, Kai and Zuo, Wangmeng and Chen, Yunjin and Meng, Deyu and Zhang, Lei},
  title   = {Beyond a Gaussian denoiser: Residual learning of deep CNN for image denoising},
  journal = {IEEE Transactions on Image Processing},
  volume  = {26},
  number  = {7},
  pages   = {3142--3155},
  year    = {2017}
}

@article{hu2018comprehensive,
  author  = {Hu, Yue and Liu, Jiahao and Liu, Xiaochuan},
  title   = {A comprehensive survey of low-rank matrix recovery for image processing},
  journal = {IEEE Access},
  volume  = {6},
  pages   = {39206--39228},
  year    = {2018}
}

@article{kolda2009tensor,
  author  = {Kolda, Tamara and Bader, Brett},
  title   = {Tensor decompositions and applications},
  journal = {SIAM Review},
  volume  = {51},
  number  = {3},
  pages   = {455--500},
  year    = {2009}
}

@article{aharon2006k,
  author  = {Aharon, Michal and Elad, Michael and Bruckstein, Alfred},
  title   = {K-SVD: An algorithm for designing overcomplete dictionaries for sparse representation},
  journal = {IEEE Transactions on Signal Processing},
  volume  = {54},
  number  = {11},
  pages   = {4311--4322},
  year    = {2006}
}

@inproceedings{buades2005non,
  author    = {Buades, Antoni and Coll, Bartomeu and Morel, Jean-Michel},
  title     = {A non-local algorithm for image denoising},
  booktitle = {CVPR},
  volume    = {2},
  pages     = {60--65},
  year      = {2005}
}

@inproceedings{vaswani2017attention,
  author    = {Vaswani, Ashish and others},
  title     = {Attention is all you need},
  booktitle = {NeurIPS},
  pages = {5998--6008},
  year      = {2017}
}

@article{dong2015image,
  author  = {Dong, Chao and Loy, Chen Change and He, Kaiming and Tang, Xiaoou},
  title   = {Image super-resolution using deep convolutional networks},
  journal = {IEEE Transactions on Pattern Analysis and Machine Intelligence},
  volume  = {38},
  number  = {2},
  pages   = {295--307},
  year    = {2016}
}

@techreport{broomhead1988radial,
  author      = {Broomhead, David and Lowe, David},
  title       = {Radial basis functions, multi-variable functional interpolation and adaptive networks},
  institution = {Royal Signals and Radar Establishment},
  year        = {1988}
}

@inproceedings{shi2016real,
  author    = {Shi, Wenzhe and Caballero, Jose and Husz{\'a}r, Ferenc and Totz, Johannes and Aitken, Andrew and Bishop, Rob and Rueckert, Daniel and Wang, Zehan},
  title     = {Real-Time Single Image and Video Super-Resolution Using an Efficient Sub-Pixel Convolutional Neural Network},
  booktitle = {CVPR},
  pages     = {1874--1883},
  year      = {2016}
}

@article{perfilieva2006fuzzy,
  author  = {Perfilieva, Irina},
  title   = {Fuzzy transforms: Theory and applications},
  journal = {Fuzzy Sets and Systems},
  volume  = {157},
  number  = {8},
  pages   = {993--1023},
  year    = {2006}
}

@article{zhang2019hyperspectral,
  author  = {Zhang, Qiang and Yuan, Qiangqiang and Li, Jie and Li, Zheng and Xu, Xiuheng and Zhang, Liangpei},
  title   = {Hybrid Noise Removal in Hyperspectral Imagery With a Spatially Adaptive Patch-Based Low-Rank Matrix Approximation},
  journal = {IEEE Transactions on Geoscience and Remote Sensing},
  volume  = {57},
  number  = {6},
  pages   = {3553--3565},
  year    = {2019}
}

@inproceedings{sainath2013low,
  author    = {Sainath, Tara N. and Kingsbury, Brian and Sindhwani, Vikas and Arisoy, Ebru and Ramabhadran, Bhuvana},
  title     = {Low-Rank Matrix Factorization for Deep Neural Network Training with High-Dimensional Output Targets},
  booktitle = {ICASSP},
  pages     = {6655--6659},
  year      = {2013}
}

@inproceedings{lebedev2014speeding,
  author    = {Lebedev, Vadim and Ganin, Yaroslav and Ranzato, Marc'Aurelio and Oseledets, Ivan and Lempitsky, Victor},
  title     = {Speeding-up Convolutional Neural Networks Using Fine-Tuned CP-Decomposition},
  booktitle = {ICLR},
  year      = {2015}
}

@article{gablonsky2000locally,
  author  = {Gablonsky, Joerg M. and Kelley, Carl T.},
  title   = {A Locally-Biased Form of the DIRECT Algorithm},
  journal = {Journal of Global Optimization},
  volume  = {21},
  number  = {1},
  pages   = {27--37},
  year    = {2001}
}

@inproceedings{chen2020dynamic,
  author    = {Chen, Yinpeng and Dai, Xiyang and Liu, Mengchen and Chen, Dongdong and Yuan, Lu and Liu, Zicheng},
  title     = {Dynamic Convolution: Attention over Convolution Kernels},
  booktitle = {CVPR},
  pages     = {11030--11039},
  year      = {2020}
}

@inproceedings{li2022omni,
  author    = {Li, Chao and Zhou, Aojun and Yao, Anbang},
  title     = {Omni-Dimensional Dynamic Convolution},
  booktitle = {ICLR},
  year      = {2022}
}

@inproceedings{dai2017deformable,
  author    = {Dai, Jifeng and Qi, Haozhi and Xiong, Yuwen and Li, Yi and Zhang, Guodong and Hu, Han and Wei, Yichen},
  title     = {Deformable Convolutional Networks},
  booktitle = {ICCV},
  pages     = {764--773},
  year      = {2017}
}

@article{liu2019multi,
  author  = {Liu, Pengju and Zhang, Hongzhi and Lian, Wei and Zuo, Wangmeng},
  title   = {Multi-Level Wavelet Convolutional Neural Networks},
  journal = {IEEE Access},
  volume  = {7},
  pages   = {74973--74985},
  year    = {2019}
}

@article{rakheja2017image,
  author  = {Rakheja, Prachi and Vig, Rekha},
  title   = {Image De-Noising Using a Combination of Wavelet Transform and Median Filtering},
  journal = {International Journal of Signal Processing, Image Processing and Pattern Recognition},
  volume  = {10},
  number  = {8},
  pages   = {41--50},
  year    = {2017}
}

@article{kutyniok2016shearlab,
  author  = {Kutyniok, Gitta and Lim, Wang-Q and Reisenhofer, Rafael},
  title   = {ShearLab 3D: Faithful Digital Shearlet Transforms Based on Compactly Supported Shearlets},
  journal = {ACM Transactions on Mathematical Software},
  volume  = {42},
  number  = {1},
  pages   = {1--42},
  year    = {2016}
}

@inproceedings{wang2018non,
  author    = {Wang, Xiaolong and Girshick, Ross and Gupta, Abhinav and He, Kaiming},
  title     = {Non-Local Neural Networks},
  booktitle = {CVPR},
  pages     = {7794--7803},
  year      = {2018}
}

@inproceedings{mairal2014convolutional,
  author    = {Mairal, Julien and Koniusz, Piotr and Harchaoui, Zaid and Schmid, Cordelia},
  title     = {Convolutional Kernel Networks},
  booktitle = {NeurIPS},
  year      = {2014}
}

@book{scholkopf2002learning,
  author    = {Sch{\"o}lkopf, Bernhard and Smola, Alexander},
  title     = {Learning with Kernels},
  publisher = {MIT Press},
  year      = {2002}
}

@inproceedings{chen2020convolutional,
  author    = {Chen, Deli and Lin, Yankai and Li, Wei and Li, Peng and Zhou, Jie and Sun, Xu},
  title     = {Measuring and Relieving the Over-Smoothing Problem for Graph Neural Networks from the Topological View},
  booktitle = {AAAI},
  year      = {2020}
}

@inproceedings{liu2018intriguing,
  author    = {Liu, Rosanne and Lehman, Joel and Molino, Piero and Such, Felipe Petroski and Frank, Eric and Sergeev, Alex and Yosinski, Jason},
  title     = {An Intriguing Failing of Convolutional Neural Networks and the CoordConv Solution},
  booktitle = {NeurIPS},
  year      = {2018}
}

@inproceedings{dosovitskiy2020image,
  author = {Dosovitskiy, Alexey and Beyer, Lucas and Kolesnikov, Alexander
          and Weissenborn, Dirk and Zhai, Xiaohua and Unterthiner, Thomas
          and Dehghani, Mostafa and Minderer, Matthias and Heigold, Georg
          and Gelly, Sylvain and Uszkoreit, Jakob and Houlsby, Neil},
  title     = {An Image is Worth 16x16 Words: Transformers for Image Recognition at Scale},
  booktitle = {ICLR},
  year      = {2021}
}

@inproceedings{woo2018cbam,
  author    = {Woo, Sanghyun and Park, Jongchan and Lee, Joon-Young and Kweon, In So},
  title     = {CBAM: Convolutional Block Attention Module},
  booktitle = {ECCV},
  pages     = {3--19},
  year      = {2018}
}

@inproceedings{wang2020eca,
  author    = {Wang, Qilong and Wu, Banggu and Zhu, Pengfei and Li, Peihua and Zuo, Wangmeng and Hu, Qinghua},
  title     = {ECA-Net: Efficient Channel Attention for Deep Convolutional Neural Networks},
  booktitle = {CVPR},
  year      = {2020}
}

@inproceedings{chen2022simple,
  author    = {Chen, Liangyu and Chu, Xiaojie and Zhang, Xiangyu and Sun, Jian},
  title     = {Simple Baselines for Image Restoration},
  booktitle = {ECCV},
  pages     = {17--33},
  year      = {2022}
}

@inproceedings{katharopoulos2020transformers,
  author    = {Katharopoulos, Angelos and Vyas, Apoorv and Pappas, Nikolaos and Fleuret, Fran{\c c}ois},
  title     = {Transformers are RNNs: Fast Autoregressive Transformers with Linear Attention},
  booktitle = {ICML},
  pages     = {5156--5165},
  year      = {2020}
}

@article{zoph2016neural,
  author  = {Zoph, Barret and Le, Quoc},
  title   = {Neural Architecture Search with Reinforcement Learning},
  journal = {arXiv preprint arXiv:1611.01578},
  year    = {2016}
}

@article{cammarasana2021kernel,
  title={Kernel-based sampling of arbitrary signals},
  author={Cammarasana, Simone and Patan{\`e}, Giuseppe},
  journal={Computer-Aided Design},
  volume={141},
  pages={103103},
  year={2021},
  publisher={Elsevier}
}

\end{document}